\documentclass[]{article}
\usepackage[letterpaper]{geometry}
\usepackage{mtsummit2017}
\usepackage{times}
\usepackage{url}
\usepackage{latexsym}
\usepackage{layout}
\usepackage{graphicx}
\usepackage{float}
\usepackage{placeins}
\usepackage{dblfloatfix}
\usepackage{wrapfig}
\usepackage[dvipsnames]{xcolor}
\usepackage{natbib}
\usepackage{apalike}
\usepackage[utf8]{inputenc}

\begin{document}


\title{Confidence through Attention}

\author{\name{\bf Matīss Rikters} \hfill  \addr{matiss@lielakeda.lv}\\ 
		\addr{Faculty of Computing, University of Latvia}\\
\AND
        \name{\bf Mark Fishel} \hfill \addr{fishel@ut.ee}\\ 
        \addr{Institute of Computer Science, University of Tartu, Estonia}
}

\maketitle
\pagestyle{empty}

\begin{abstract}
Attention distributions of the generated translations are a useful bi-product of attention-based recurrent neural network translation models and can be treated as soft alignments between the input and output tokens. In this work, we use attention distributions as a confidence metric for output translations. We present two strategies of using the attention distributions: filtering out bad translations from a large back-translated corpus, and selecting the best translation in a hybrid setup of two different translation systems. While manual evaluation indicated only a weak correlation between our confidence score and human judgments, the use-cases showed improvements of up to 2.22 BLEU points for filtering and 0.99 points for hybrid translation, tested on English\(\leftrightarrow\)German and English\(\leftrightarrow\)Latvian translation.
\end{abstract}

\section{Introduction}

Neural machine translation (NMT) has recently redefined the state-of-the-art in machine translation \citep{sennrich2016,wu2016google}, with one of the ground-breaking innovations that enabled this being the introduction of the attention mechanism \citep{DBLP:journals/corr/BahdanauCB14}. It enables the model to find parts of a source sentence that are relevant to predicting a target word (pay attention), without the need to form these parts as a hard segment explicitly. Decoding sentences with the attention-based model resulted in a useful by-product -- soft alignments between tokens of source and target sentences.
These can be used for many purposes, such as replacing unknown words with back-off translations from a dictionary \citep{jean-EtAl:2015:ACL-IJCNLP} and visualizing the soft alignments \citep{pbmlnmtvis}.

In this paper, we propose using the attention alignments as an indicator of the translation output quality and the confidence of the decoder. We define metrics of confidence that detect and penalize under-translation and over-translation \citep{modelling-coverage} as well as input and output tokens with no clear alignment, assuming that all these cases most likely mean that the quality of the translation output is bad.

We apply these attention-based metrics to two use-cases: scoring translations of an NMT system and filtering out the seemingly unsuccessful ones, and comparing translations from two different NMT systems, in order to select the best one.

The structure of this paper is as follows: Section~\ref{sctRelWork} summarizes related work in back-translating with NMT, machine translation combination approaches and confidence estimation. Section~\ref{sctScoring} introduces the problem of faulty attention distributions and a way to quantify it as a confidence score. Sections \ref{sctFiltering} and \ref{sctHybrid} outline the two use-cases for this score -- translation filtering and hybrid selections. Finally, we conclude in Section \ref{sctConclusions} and mention directions for future work in Section \ref{sctFuture}.

\section{Related Work}
\label{sctRelWork}

\subsubsection*{Back-translation of Monolingual Data}

One of the first uses of back-translation of monolingual data as an additional source of training data was reported by \citep{sennrich2016} in their submission for the WMT16 news translation shared task. They translated target-language monolingual corpora into the source language of the respective language pair, and then used the resulting synthetic parallel corpus as additional training data. They performed experiments in ranges from 2 million to 10 million back-translated sentences and reported an increase of 2.2 - 7.7 BLEU \citep{papineni2002bleu} for translating between English and Czech, German, Romanian and Russian. The authors also experimented with different amounts of back-translated data and found that adding more data gradually improves performance.

In a later paper \citet{sennrich-haddow-birch:2016:P16-11} explored other methods of using monolingual data. They experimented with adding an enormous amount of monolingual sentences as targets without any sources to the parallel corpus and compared that to performing back-translation on a part of the monolingual data. While both methods outperform using just parallel data, the back-translated synthetic parallel corpus is a much more powerful addition than the mono data alone.

\citet{pinnis2017neural} experimented with using large and even larger amounts of back-translated data and came to a conclusion that any amount is an improvement, but using double the amount gives lower results, while still better than not using any at all. These results hint that it may be possible to get even better results when using only the part of the data selected with some criterion. One of the aims of our work is to provide one such criterion.

\subsubsection*{Machine Translation System Combination}

\citet{Zhou2017} used attention to combine outputs from NMT and SMT systems. The authors first trained intermediate NMT, SMT and hierarchical SMT systems with one-half of the training data. Afterwards, they used each system to translate the target side of the other half of the training data. Finally, the three translated parts as source sentence variants along side the clean target sentence were used for training the combination neural network. This approach gave the network more choices of where to pay attention and which parts should be ignored in the training process. They perform experiments on Chinese\(\rightarrow\)English and report BLEU score improvement by 5.3 points over the best single system and 3.4 points over traditional MT combination methods.

\citet{Peter2016} perform MT system combination in a more traditional manner - using confusion networks. They use 12 different SMT and NMT systems to generate hypothesis translations, align and reorder each hypothesis to match one skeleton hypothesis, creating a confusion network. For the final output is generated by finding the best path in the network. The authors report an improvement of 1.0 BLEU compared to the best single system, translating from English into Romanian.

\subsubsection*{Translation Confidence Metrics}

Lately the idea of modeling coverage in NMT was introduced, for example, \citet{modelling-coverage} integrate it directly into the attention mechanism and report improved translation quality as a result. On the simpler side of things, \citet{bridgingGoogle} perform tests with a baseline attention that uses an additional coverage penalty at decoding time; they report no improvement compared to the common length normalization. Our metrics are partially motivated by the coverage penalty, though we apply them at the post-translation stage to determine the confidence of the decoder and the quality of the already made translation, which makes it applicable regardless of which software or approach were used.

Another closely related task is quality estimation. The dominating approach there is collecting post-edits and training a machine learning model to predict the quality score or classify translations into usable/not, near-perfect/not, etc \citep{quality-estimation, quality-estimation-ml}. The main similarity between our work and quality estimation is their usage of glass-box features (i.e. information about the MT system or the decoder's internal parameters). While our approach does not cover all aspects of quality estimation, it requires no data or training and can be applied to any language and neural machine translation system.




\section{Penalizing Attention Disorders}
\label{sctScoring}

Before describing the confidence metrics based on attention weights, here is a brief overview of the NMT architecture where the attention weights come from.

\subsection{Source of Attention}

Our work is built around the encoder-decoder machine translation approach \citep{sutskever2014sequence,cho-EtAl:2014:EMNLP2014} with an attention mechanism \citep{DBLP:journals/corr/BahdanauCB14}. In this approach the source tokens are learned to be represented by an encoder, which consists of an embedding layer and a bi-directional LSTM or GRU layer \citep[or 8,][]{bridgingGoogle}, the outputs of which serve as the learned representation.

There is also a decoder that consists of another layer (or 8, ibid.) of LSTM/GRU cells, with an output layer for predicting the softmax-encoded raw probability distribution of each output word, one at a time. The state of the decoder layer(s) and thus the output distribution depends on the previous recurrent states, the previously produced output word and a weighted sum of the representations of the source sentence tokens. The weights in this sum are generated for every output word by the attention mechanism, which is a feed-forward neural network with the previous state of the decoder and each input word representation as input and the raw weight of that word for the next state as output. Finally, the attention weights are normalized as follows:
\[
\alpha_{ij} = \frac{\exp(e_{ij})}{\sum_k \exp(e_{ik})}
\]
where $e_{ij}$ is the raw predicted weight and $\alpha_{ij}$ -- the final attention weight between the input token $j$ and output token $i$.

Once the encoder-decoder network has been trained, it can be used to produce translations by predicting the probability for each next word, which can serve as the basis for sampling, greedy search or beam search \citep{sennrich2017nematus}. We refer the reader for a complete description to \citep{DBLP:journals/corr/BahdanauCB14} and ourselves turn on to the main topic of the paper that uses the weights $\alpha_{ij}$ to estimate the confidence of the translations.

Together with the translation, it is also possible to save the attention values between the input tokens and each produced output token. These values can be interpreted as the influence of the input token on the output token, or the strength of the connection between them. Thus, weak or dispersed connections should intuitively indicate a translation with low confidence, while high values and strong connections between one or two tokens on both sides should indicate higher confidence. Next, we present our take at formalizing this intuition.

\subsection{Measuring Attention}

\begin{figure*}[t]
  \includegraphics[width=\linewidth]{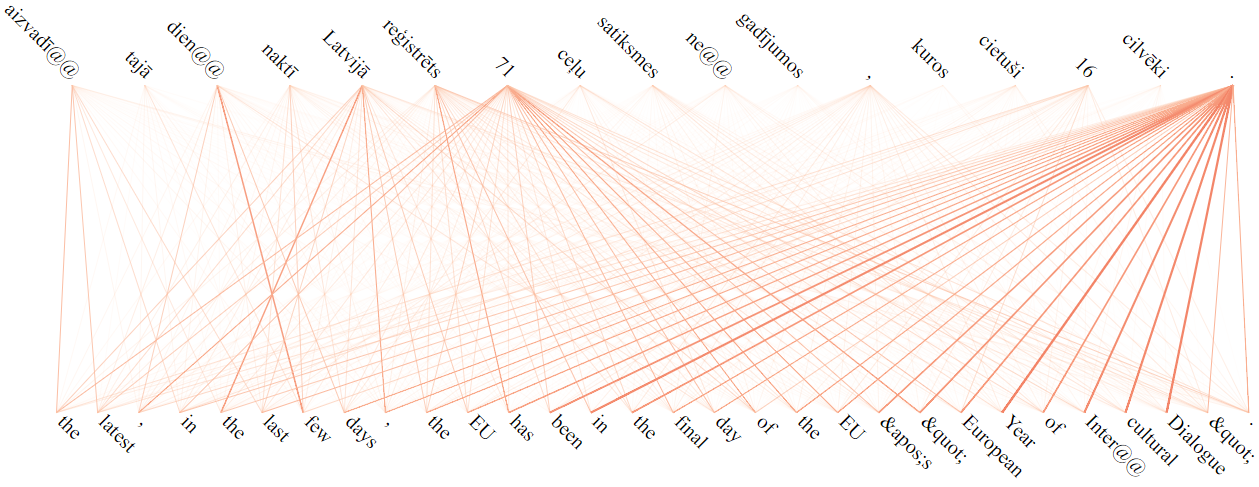}
  \caption{Attention alignment visualization of a bad translation. Reference translation: \emph{71 traffic accidents in which 16 persons were injured have happened in Latvia during the last 24 hours.}, hypothesis translation: \emph{the latest , in the last few days , the EU has been in the final day of the EU 's " European Year of Intercultural Dialogue "}. $CDP = -0.900$, $AP_{out} = -2.809$, $AP_{in} = -2.137$, $Total = -5.846$.}
  \label{fig:bad-attention}
\end{figure*}

\begin{figure*}[t]
  \includegraphics[width=\linewidth]{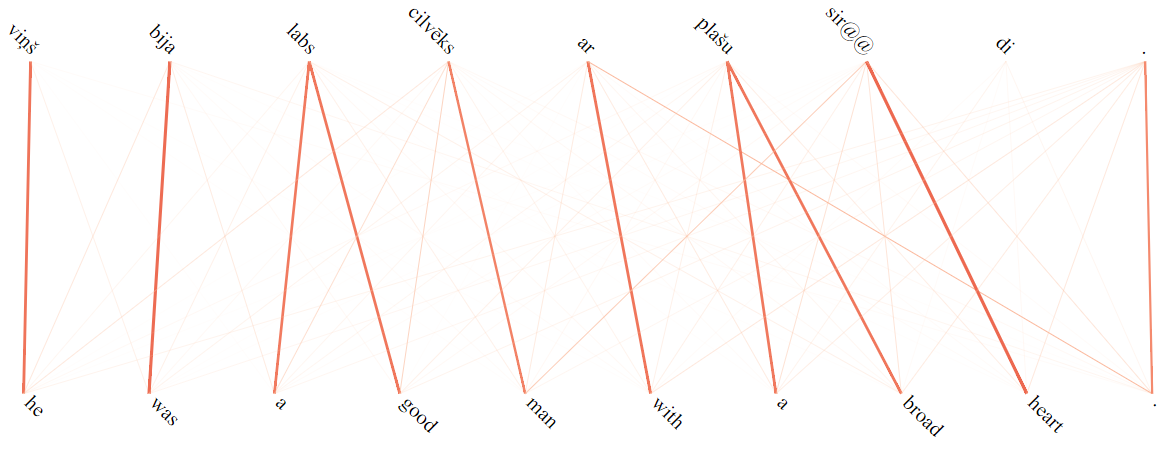}
  \caption{Attention alignment visualization of a good translation. Reference translation: \emph{He was a kind spirit with a big heart.}, hypothesis translation: \emph{he was a good man with a broad heart}. $CDP = -0.099$, $AP_{out} = -1.077$, $AP_{in} = -0.847$, $Total = -2.024$.}
  \label{fig:good-attention}
\end{figure*}

Figure~\ref{fig:bad-attention} shows an example of a translation that has little or nothing to do with the input, a frequent occurrence in NMT. Besides the text of the translation, it is clear already by looking at the attention weights of this pair that the translation is weak:
\begin{itemize}
	\item some input tokens (like the sentence-final full-stop) are most strongly connected to several unrelated output tokens, in other words, their coverage is too high,
    \item most of the input token attentions, as well as some output token attentions, are highly dispersed, without one or two clear associations on the counterpart.
\end{itemize}
On the other hand, a picture like Figure~\ref{fig:good-attention} intuitively corresponds to a good translation, with strongly focused alignments. It is this intuition that our metrics formalize: penalizing translations with tokens with a total coverage of not just below but much higher than 1.0, as well as tokens with a dispersed attention distribution.

\subsubsection*{Coverage Deviation Penalty}

Previous work \citep{bridgingGoogle} defines a coverage penalty, which is meant to punish translations for not paying enough attention to input tokens:
\[
CP = \beta \sum_j \log ( \min ( \sum_i \alpha_{ji} , 1.0 ) ),
\]
where $i$ is the output token index, $j$ -- the input token index, $\beta$ is used to control the influence of the metric and $CP$ -- the coverage penalty.

The first part of our metric draws inspiration from the coverage penalty; however, it penalizes not just lacking attention but also too much attention per input token. The aim is to penalize the sum of attentions per input token for going too far from 1.0\footnote{This could be replaced with the token's expected fertility, which we leave for future work}, so tokens with total attention of 1.0 should get a score of 0.0 on the logarithmic scale, while tokens with less attention (like 0.2) or more attention (like 2.5) should get lower values. We thus define the coverage deviation penalty:
\[
CDP = -\frac{1}{J} \sum_j \log \left(1 + (1 - \sum_i \alpha_{ji} )^2 \right),
\]
where $J$ is the length of the input sentence. The metric is on a logarithmic scale, and it is normalized by the length of the input sentence in order to avoid assigning higher scores to shorter sentences\footnote{This is not required for choosing translations of the same sentence by the same system, but is required in our experiments described in the next sections.}. See examples of the CDP metric's values on Figures~\ref{fig:bad-attention} and \ref{fig:good-attention}.


\subsubsection*{Absentmindedness Penalty}

However, it is not enough to simply cover the input, we conjecture that more confident output tokens will allocate most of their attention probability mass to one or a small number of input tokens. Thus the second part of our metric is called the absentmindedness penalty and targets scattered attention per output token, where the dispersion is evaluated via the entropy of the predicted attention distribution. Again, we want the penalty value to be 1.0 for the lowest entropy and head towards 0.0 for higher entropies.
\[
AP_{out} = -\frac{1}{I} \sum_i \sum_j \alpha_{ji} \cdot \log \alpha_{ji}
\]
The values are again on the log-scale and normalized by the source sentence length $I$.

The absentmindedness penalty can also be applied to the input tokens after normalizing the distribution of attention per input token, resulting in the counter-part metric $AP_{in}$. This is based on the assumption that it is not enough to cover the input token, but rather the input token should be used to produce a small number of outputs. See examples of both metric's values on Figures~\ref{fig:bad-attention} and \ref{fig:good-attention}.

Finally, we combine the coverage deviation penalty with both the input and output absentmindedness penalties into a joint metric via summation:
\[
confidence = CDP + AP_{out} + AP_{in}
\]

Next, we evaluate the metrics directly against human judgments and indirectly by applying them to filtering translations and plugging them into a sentence-level hybrid translation scheme.

\subsection{Human Evaluation}

It is clear that the defined metrics only paint a partial picture, since they rely on the attention weights only. For instance, they do not evaluate the lexical correspondence between the source and hypothesis, and more generally, being confident does not mean being right. We wanted to find out how much confidence in our case correlates with translation quality.

To do so we asked human volunteers to perform pairwise ranking of translations from two baseline NMT systems: one done with Nematus \citep{sennrich2017nematus} and the other -- with Neural Monkey \citep{helcl2017neural}. The translations and measurements were done for English-Latvian and Latvian-English, using corpora from the news translation shared task of WMT'2017; further details can be found in Section~\ref{sctFiltering}. We selected 200 random sentences for both translation directions and these were given to native Latvian speakers for evaluation. The MT-EQuAl \citep{Girardi2014} tool was used for the evaluation task. The evaluators were shown one source sentence at a time along with the two different translations. They were instructed to assign one of five categories for each translation: "worst", "bad", "ok", "good" or "best", noting that both may be categorized as equally "good" or "bad", etc. Differing judgments for the same sentence were averaged. All 200 sentences were annotated by at least one human annotator.

It makes more sense to treat the results as relative comparisons, not absolute scores, as the annotators only see two translations at a time. We use these comparisons to compute the Kendall rank correlation coefficient \citep{kendallTau} by only looking at the pairs where human scores differ. Since we only have comparisons for each pair and not between different sentences, the coefficient is computed as
\[
\tau = \frac{pos - neg}{pos + neg},
\]
where $pos$ is the number of pairs where the metric agrees with the human judgment and $neg$ is the number of pairs where they disagree.

The results are presented in Table~\ref{tblCorrelations}, and as we can see they indicate weak correlation, with the absolute values of $\tau$ between $0.012$ and $0.200$.
\begin{table}[h]
\begin{center}
\begin{tabular}{|l|l|l|l|l|}
\hline
Language pair & CDP & AP$_{in}$ & AP$_{out}$ & Overall \\
\hline
En\(\rightarrow\)Lv & 0.099 & 0.074 & 0.123 & 0.086 \\
Lv\(\rightarrow\)En & -0.012 & -0.153 & -0.200 & -0.153 \\
\hline
\end{tabular}
\end{center}
\caption{The Kendall's Tau correlation between human judgments and the confidence scores.}
\label{tblCorrelations}
\end{table}

\begin{figure*}[h]
  \includegraphics[width=\linewidth]{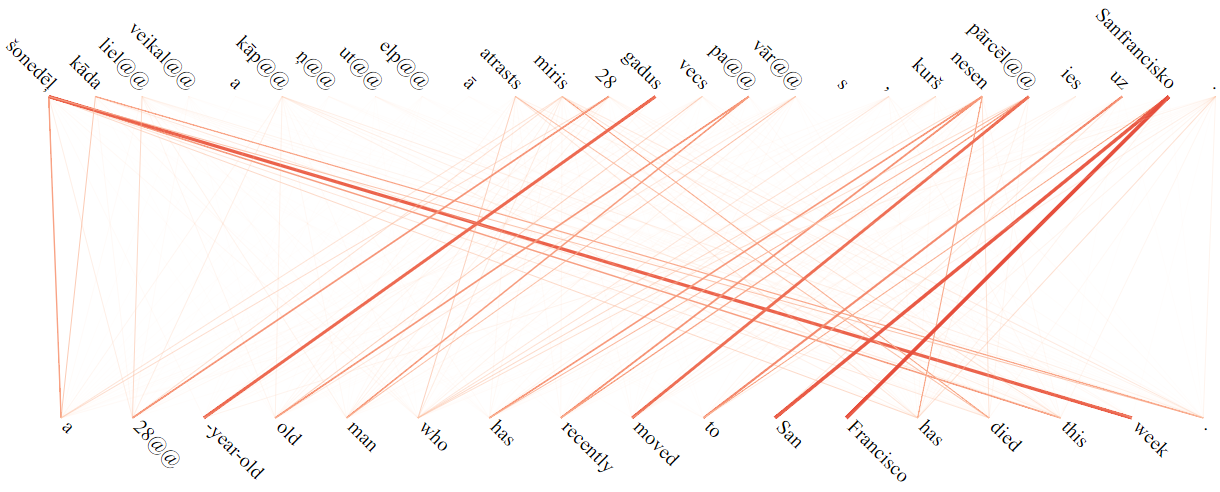}
  \caption{Attention alignment visualization of a bad translation. Reference translation: \emph{a 28-year-old chef who had recently moved to San Francisco was found dead in the stairwell of a local mall this week .}, hypothesis translation: \emph{a 28-year-old old man who has recently moved to San Francisco has died this week .}, $CDP = -0.250$, $AP_{out} = -1.740$, $AP_{in} = -1.46$, $Total = -3.45$.}
  \label{fig:bad-attention-good-translation}
\end{figure*}

Let us look closer at where the metrics disagree with human judgments. Figure \ref{fig:bad-attention-good-translation} shows an example of a translation which was rated highly by human annotators but poorly with our metrics. While the sentence is a good translation, it does not follow the source word-by-word. Some subword units and functional words do not have a clear alignment, even though they are understood/generated correctly. This means that one problem with our metrics is that they might be over-penalizing translations that deviate from a direct literal translation.

Next, we continue with the experiments of using our metrics to filter synthetic data and to select translations in a hybrid MT scenario.

\FloatBarrier

\section{Filtering Back-translated Data}
\label{sctFiltering}
\subsection{Baseline Systems and Data}

Our baseline systems were trained with two NMT frameworks - Nematus (NT) \citep{sennrich2017nematus} and Neural Monkey (NM) \citep{helcl2017neural}. 
For all NMT models we used a shared subword unit vocabulary \citep{SennrichBPE2016} of 35000 tokens, clip the gradient norm to 1.0 \citep{pascanu2013difficulty}, dropout of 0.2, trained the models with Adadelta \citep{zeiler2012adadelta} and performed early stopping after 7 days of training. For models with each NMT framework we used the default settings as mentioned in the frameworks documentation:

\begin{itemize}
\item For NT models we used a maximum sentence length of 50, word embeddings of size 512, and hidden layers of size 1000. For decoding with NT we used beam search with a beam size of 12.

\item For NM models we used a maximum sentence length of 70, word embeddings and hidden layers of size 600. For decoding with NM a greedy decoder was used.
\end{itemize}

Training, development and test data for all systems in both language pairs and translation directions was used from the WMT17 news translation task \footnote{EMNLP 2017 Second Conference on Machine Translation - http://www.statmt.org/wmt17/}. For the baseline systems, we used all available parallel data, which is 5.8 million sentences for En\(\leftrightarrow\)De and 4.5 million sentences for En\(\leftrightarrow\)Lv.  

\subsection{Back-translating and Filtering}
We used our baseline En\(\rightarrow\)Lv and Lv\(\rightarrow\)En NM and NT systems to translate all available Latvian monolingual news domain data - 6.3 million sentences in total from \textit{News Crawl: articles from 2014, 2015, 2016}, and the first 6 million sentences from the English \textit{News Crawl 2016}. Much more monolingual data was available from other domains aside from news. Since the development and test data was of the news domain, we only used that, considering it as in-domain data for our systems.

For each translation, we used the attention provided from the NMT system to calculate our confidence score, sorted all translations according to the score and selected the top half of the translations along with the corresponding source sentences as the synthetic parallel corpus. We used only the full confidence score (combination of CDP, AP$_{out}$ and AP$_{in}$) for filtering instead of each individual score due to its smoother overall correlation with human judgments. In between, we also removed any translation that contained any \textit{\(<\)unk\(>\)} tokens. 

To compare attention-based filtering with a different method, we trained a CharRNN\footnote{Multi-layer Recurrent Neural Networks (LSTM, GRU, RNN) for character - level language models in Torch https://github.com/karpathy/char-rnn} language model (LM) with 4 million news sentences from each of the target languages. We used these LMs to get perplexity scores for all translations, order them and get the \textit{better half}. Table \ref{tab:human-fil-results} summarizes how much human evaluation overlaps with each of the filtering methods. The final row indicates how much both filtering methods overlap with each other. While results from either approach don't look overly convincing, the LM-based approach has been proven to correlate with human judgments close to the BLEU score and is a good evaluation method for MT without reference translations \citep{gamon2005sentence}. Therefore the attention-based approach that does not require training of an additional model and overlaps with human judgments to approximately the same level should be more desirable.  

\begin{table}[h]
  \begin{center}
    \begin{tabular}{|l|c|c|}
      \hline 
      Filtering Method & En\(\rightarrow\)Lv & Lv\(\rightarrow\)En\\ 
      \hline 
      LM-based overlap with human 				& 58\% & 56\% \\
      Attention-based overlap with human 		& 52\% & 60\% \\
      \hline
      LM-based overlap with Attention-based 	& 34\% & 22\% \\
      \hline
    \end{tabular}
  \end{center}
  \caption{Human judgment overlap results on 200 random sentences from the \textit{newsdev2017} dataset compared to filtering methods.}
  \label{tab:human-fil-results}
\end{table}

\subsection{NMT with Filtered Synthetic Data}

\begin{figure*}[h]
  \includegraphics[width=\linewidth]{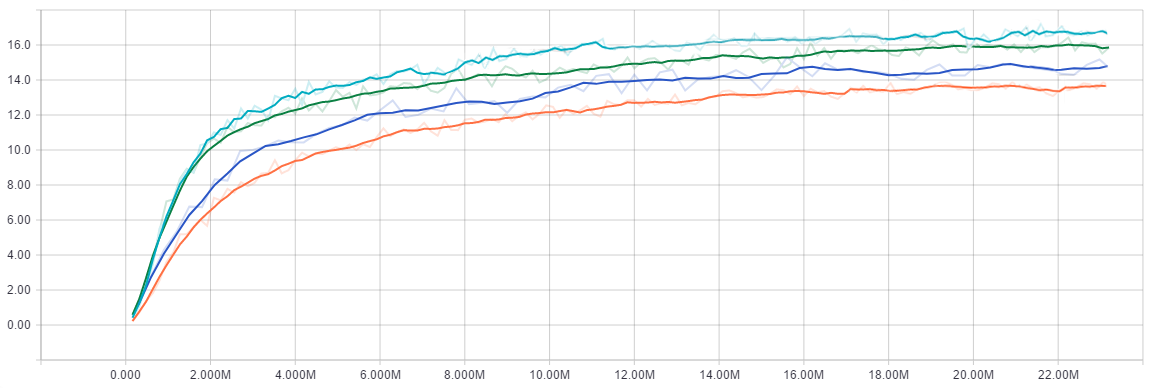}
  \caption{Automatic evaluation progression of Lv\(\rightarrow\)En experiments on validation data. Orange -- baseline; dark blue –- with full back-translated data; green -- with LM-filtered back-translated data; light blue -- with attention-filtered back-translated data.}
  \label{fig:bleu-progression}
\end{figure*}

We shuffled each synthetic parallel corpus with the baseline parallel corpora and used them to train NMT systems. In addition to the baseline and two types of filtered BT synthetic data, we also trained a system with the full BT data for each translation direction. Figure \ref{fig:bleu-progression} shows a combined training progress chart for Lv\(\rightarrow\)En on the full \textit{newsdev2017} dataset that was used as the development set for training. Here the differences between all four approaches are clearly visible. Further results on a subset of \textit{newsdev2017} and the full \textit{newstest2017} dataset are summarized in Table \ref{tab:lv-fil-results}. While for Lv\(\rightarrow\)En and En\(\leftrightarrow\)De the attention-based approach is the clear leader, for En\(\rightarrow\)Lv it falls behind the LM filtered version. We were not able to identify a clear reason for this and leave it for the future work. As expected, adding BT synthetic training data allows to get higher BLEU scores in all cases. It can be observed that filtering out half of the badly translated data and keeping only the best translations either does not decrease the final output quality in some cases or even further increase the quality in others, when using the LM. With filtering by attention, the results are more inconsistent - even higher in one direction while deterioration in the other. A reason for this could be that for Lv\(\rightarrow\)En attention-based filtering the similarity with human judgments was higher than for En\(\rightarrow\)Lv (Table \ref{tab:human-fil-results}), and it was also more different from the LM-based one. While for the other direction it is the other way around. 

\begin{table}[h]
  \begin{center}
    \begin{tabular}{|l|c|c|c|c|c|c|c|c|}
      \hline 
       & \multicolumn{8}{|c|}{BLEU} \\ 
      \hline 
      Dataset & Dev & Test  & Dev & Test & Dev & Test  & Dev & Test \\ 
      \hline 
      System & \multicolumn{2}{|c|}{En\(\rightarrow\)Lv}  & \multicolumn{2}{|c|}{Lv\(\rightarrow\)En} & \multicolumn{2}{|c|}{En\(\rightarrow\)De}  & \multicolumn{2}{|c|}{De\(\rightarrow\)En} \\ 
      \hline
      Baseline 							& 8.36 	& 11.90 	& 8.64 		& 12.40 	& 25.84 & 20.11 & 30.18 & 26.26 \\
      + Full Synthetic 					& 9.42 	& 13.50 	& 9.01 		& 13.81 	& 28.97 & 22.68 & 34.82 & 29.35 \\
      + LM-Filtered Synthetic 			& 9.75 	& 13.52 	& 9.45 		& 14.30 	& 29.59 & 23.48 & 34.47 & 29.42 \\
      \bf + Attn.-Filtered Synth. 		& 8.99 	& 12.76 	& \bf11.23 	& \bf14.83 	& \bf30.19 & 23.16 & \bf35.19 & \bf29.47 \\
      \hline
    \end{tabular}
  \end{center}
  \caption{Experiment results in BLEU for translating between English\(\leftrightarrow\)Latvian with different types of back-translated data using development (200 random sentences from \textit{newsdev2017}) and test (\textit{newstest2017}) datasets.}
  \label{tab:lv-fil-results}
\end{table}

\FloatBarrier

\section{Attention-based Hybrid Decisions}
\label{sctHybrid}

We translated the development set with both baseline systems for each language pair in each direction. The hybrid selection of the best translation was performed similarly to filtering, where we discarded the worst-scoring half of the translations. In the hybrid selection, we used the same score to compare both translations of a source sentence and choose the better one. Results of the hybrid selection experiments are summarized in Table \ref{tab:hyb-results}. For translating between En\(\leftrightarrow\)Lv, where the difference between the baseline systems is not that high (0.06 and 1.55 BLEU), the hybrid method achieves some meaningful improvements. However, for En\(\leftrightarrow\)De, where differences between the baseline systems are bigger (3.46 and 4.46 BLEU), the hybrid drags both scores down.  

\begin{table}[h]
  \begin{center}
    \begin{tabular}{|l|c|c|c|c|}
      \hline 
       & \multicolumn{4}{|c|}{BLEU} \\ 
      \hline 
      System & En\(\rightarrow\)De & De\(\rightarrow\)En & En\(\rightarrow\)Lv & Lv\(\rightarrow\)En\\ 
      \hline 
      Neural Monkey	& 18.89 	& 26.07 	& 13.74 		& 11.09 \\
      Nematus		& 22.35 	& 30.53 	& 13.80 		& 12.64 \\
      \bf Hybrid 	& 20.19 	& 27.06 	& \bf 14.79 	& \bf 12.65 \\
      \hline 
      Human 		& 23.86 	& 34.26 	& 15.12 		& 13.24 \\
      \hline
    \end{tabular}
  \end{center}
  \caption{Hybrid selection experiment results in BLEU on the development dataset (200 random sentences from \textit{newsdev2017}).}
  \label{tab:hyb-results}
\end{table}

The last row of the results Table \ref{tab:hyb-results} shows BLEU scores for the scenario when human annotator preferences were used to select each output sentence. An overview of human evaluator preferred translation selections is visible in Table \ref{tab:human-hyb-results}. The results show that out of all translations the human evaluators deliberately prefer one or the other system. Aside from En\(\rightarrow\)Lv, where a slight tendency towards Neural Monkey translations can be observed, all others look more or less equal. This highly contrasts with the BLEU scores from Table \ref{tab:hyb-results}, where in both translation directions from English human evaluators prefer the lower-scoring system more often than the higher-scoring one.
The final row of Table \ref{tab:human-hyb-results} shows how much our attention-based score matches the human judgments in selecting the best translation.

\begin{table}[h]
  \begin{center}
    \begin{tabular}{|l|c|c|c|c|}
      \hline 
      System & En\(\rightarrow\)De  & De\(\rightarrow\)En & En\(\rightarrow\)Lv &  Lv\(\rightarrow\)En\\ 
      \hline 
      Neural Monkey						& 54\% 		& 42\% 		& 61.5\% 	& 47\% \\
      Nematus							& 46\% 		& 58\% 		& 38.5\% 	& 53\% \\
      \hline 
      Overlaps with hybrid selection 	& 57\% 		& 47\% 		& 62.5\% 	& 51\% \\
      \hline
    \end{tabular}
  \end{center}
  \caption{Human evaluation results on 200 random sentences from the \textit{newsdev2017} dataset compared to attention-hybrid selection.}
  \label{tab:human-hyb-results}
\end{table}

\FloatBarrier

\section{Conclusions}
\label{sctConclusions}

In this paper, we described how attentional data from neural machine translation systems can be useful for more than just visualizations or replacing specific tokens in the output. We introduced an attention-based confidence score that can be used for judging NMT output. Two applications of using attentional data were investigated and compared to similar approaches. We used a smaller dataset to perform manual evaluation and compared that to all automatically obtained results. Our experiments showed interesting results and some increases in automated evaluation, as well as a good correlation with human judgments.

In addition to the methods described in this paper, we release open-source scripts\footnote{Confidence Through Attention - https://github.com/M4t1ss/ConfidenceThroughAttention} for (1) scoring, ordering and filtering NMT translations, (2) performing hybrid selections between two different NMT outputs of the same source, and (3) software for inspecting attention alignments that the NMT systems produce in the translation process (used for Figures \ref{fig:bad-attention} and \ref{fig:good-attention}). We also provide all development subsets that we used for manual evaluation with anonymized human annotations.

\section{Future Work}
\label{sctFuture}

This paper introduced the first steps in using NMT attention for less obvious intentions. It seemed that the attention score can complement the LM perplexity score in distinguishing good from bad translations. An idea for future experiments could be combining these scores to achieve a higher correlation with human judgments.

Additional improvements can be made to the hybrid decisions as well. Since the score represents the systems \textit{confidence}, a badly trained NMT system can be more confident about a bad translation than a good system about a decent translation. While a hybrid combination of two similar quality NMT systems did put the attention score to good use, in the case with different quality systems the confidence of the weaker one was a pitfall. This indicates that the confidence score could be used in ensemble with a quality estimation score or used as a feature in training an MT quality estimation system.

For filtering synthetic back-translated data we dropped the worst-scoring 50\% of the data, but this threshold may not be optimal for all scenarios. Several paths worth more exploration include exploring the effects of different static thresholds (e.g. 30\% or 70\%) or clustering the data by confidence score and dropping the lowest-scoring one or two clusters. Another path worth exploring for filtering would be to see how filtering by each individual score (CDP, $AP_{in}$, $AP_{out}$) compares to filtering by confidence. 

In the near future, we also plan to supplement an attention inspection tool so that it displays confidence metrics and additional visualizations based on these scores.

\FloatBarrier

\bibliographystyle{apalike}
\bibliography{mtsummit2017}

\end{document}